\pdfoutput=1
\documentclass[11pt]{article}

\usepackage{soul}
\usepackage{graphicx}
\usepackage{booktabs}
\usepackage{multirow}  
\usepackage{xcolor}
\usepackage{newfloat}
\usepackage{listings}
\usepackage{enumitem}
\usepackage{balance}
\usepackage{colortbl}
\usepackage{algpseudocode}
\usepackage{algorithm} 
\usepackage{wrapfig}
\usepackage{amssymb}
\usepackage{amsmath}
\newcommand{\aaa}[1]{{\color{magenta}{#1}}}
\newcommand{\myPara}[1]{\vspace{.1in} \noindent\textbf{#1}}
\usepackage[final]{acl}

\usepackage{times}
\usepackage{latexsym}

\usepackage[T1]{fontenc}
\usepackage[utf8]{inputenc}

\usepackage{microtype}

\usepackage{inconsolata}

\usepackage{graphicx}

\title{MoExtend: Tuning New Experts for Modality and Task  Extension}

\author{
 \textbf{Shanshan Zhong\textsuperscript{1}},
 \textbf{Shanghua Gao\textsuperscript{2}},
 \textbf{Zhongzhan Huang\textsuperscript{1}},
 \textbf{Wushao Wen\textsuperscript{1}},
\\
 \textbf{Marinka Zitnik\textsuperscript{2}},
 \textbf{Pan Zhou\textsuperscript{3}}
\\
 \textsuperscript{1}Sun Yat-sen University,
 \textsuperscript{2}Harvard University,
 \textsuperscript{3}Singapore Management University,
\\
 \small{
   \textbf{Correspondence:} \href{mailto:panzhou@smu.edu.sg}{panzhou@smu.edu.sg}
 }
}

\begin{document}
\maketitle
\begin{abstract}
Large language models (LLMs) excel in various tasks but are primarily trained on text data, limiting their application scope. Expanding LLM capabilities to include vision-language understanding is vital, yet training them on multimodal data from scratch is challenging and costly. Existing instruction tuning methods, e.g., LLAVA, often connects  a pretrained CLIP vision encoder and LLMs via fully fine-tuning LLMs to  bridge the modality gap. However, full fine-tuning is plagued by catastrophic forgetting, i.e., forgetting previous knowledge,  and high training costs particularly in the era of increasing tasks and modalities.  
To solve this issue, we introduce MoExtend, an effective framework designed to streamline the modality adaptation and extension of Mixture-of-Experts (MoE) models. MoExtend seamlessly integrates new experts into pre-trained MoE models, endowing them with novel knowledge without the need to tune pretrained models such as MoE and vision encoders. This approach enables rapid adaptation and extension to new modal data or tasks, effectively addressing the challenge of accommodating new modalities within LLMs. Furthermore, MoExtend avoids tuning pretrained models, thus mitigating the risk of catastrophic forgetting. Experimental results demonstrate the efficacy and efficiency of MoExtend in enhancing the multimodal capabilities of LLMs, contributing to advancements in multimodal AI research. \aaa{https://github.com/zhongshsh/MoExtend}.
\end{abstract}

\section{Introduction}
\label{sec:intro}
% \gsh{Will we working on vision and other moda for this version?}
General-purpose large language models (LLMs) have demonstrated their effectiveness across a broad spectrum of application scenarios, such as conversational chatbot~\cite{ouyang2022training}, document analysis~\cite{radford2019language}, and coding~\cite{chen2021evaluating}. 
While the most powerful LLMs, such as ChatGPT~\cite{radford2019language}, Llama~\cite{touvron2023llama}, and Mixtral~\cite{jiang2024mixtral}, are predominantly trained on textual data, there is a growing interest in extending their capabilities to support a wider array of applications beyond natural language processing, especially with a significant focus on vision-language understanding~\cite{liu2023improved,zhu2023minigpt,liu2023llava,team2023gemini}.
While training large models from scratch on multimodal data suffers from insufficient data~\cite{zhu2023minigpt} and significant training costs~\cite{team2023gemini}, 
most efforts have been focused on enhancing the multimodal capabilities of pretrained LLMs~\cite{zhu2023minigpt,liu2023llava,liu2023improved}. 
% \zp{@shanghua, please add corresponding citations here. Shanshan has no so much efforts to add them. Only one day to submit}
%
To accomplish this, 
LLMs handle new modal data by processing representations extracted by encoders specific to each modality. For instance, the vision transformer pre-trained with CLIP~\cite{radford2021learning} is utilized to encode visual images. Then, the model is trained using text-image Q\&A pairs to carry out tasks based on these multimodal instructions.

The parameter-efficient approach to bridging the gap between modality-specific encoders and large language models (LLMs) involves the use of a few linear projection layers~\cite{zhu2023minigpt} and Low-Rank Adaptation (LoRA)~\cite{zhang2023llamaadapter,hu2021lora}. However, this does not entirely mitigate the modality gap, limiting LLMs' ability to fully understand new modalities. 
Consequently, State-of-the-art multimodal methods, e.g. LLaVA~\cite{liu2023llava}, have sought to further enhance the multimodal capabilities of LLMs by fully fine-tuning these models on multimodal datasets~\cite{lin2024moe}.
Despite these efforts, fully fine-tuning encounters two significant obstacles:
1) 
\textbf{Catastrophic Forgetting}:
LLMs, when fine-tuned to effectively integrate various modalities, tend to lose the knowledge they had acquired previously~\cite{luo2023empirical}. 
2) \textbf{Large fine-tuning cost}:
With the increasing sizes of LLMs, fully fine-tuning on larger models is becoming increasingly impractical. As a result, smaller models, like those with 7 billion parameters, are often preferred. However, this preference restricts the exploration and utilization of the capabilities of larger LLMs.
How to efficiently extend new modality to large LLM while reduce the side effect of catastrophic forgetting is an urging problem for multimodal LLMs.

Mixture-of-Experts (MoE) architectures
enable LLMs to use the gate layer to dynamically select the most relevant experts from a diverse set of specialized experts, e.g. different MLP layers in Transformer, for a given query token.
MoE helps to enlarge the model size by increase the number of experts while
keeping low inference cost by selecting a sub set of experts for each token.
For instance, the Mixtral-8x7B model~\cite{jiang2024mixtral} incorporates 8 MLP experts per block, totaling 46.7 billion parameters, yet it selects only 2 experts, utilizing 12.9 billion parameters per token.
Nonetheless, the current MoE models predominantly concentrates on the textual modality.

We introduce an extension strategy for MoE models, named MoExtend, designed to accommodate new modalities. 
This strategy involves incorporating new modality-specific experts and calibration modules into trained MoE models to enhance their capability to process additional modalities.
MoExtend maintains the original MoE model parameters unchanged, while 
only trains the newly added experts and the corresponding gate layer.
By doing so,
MoExtend facilitates the efficient adaptation of new modalities into large models while also addressing issues of catastrophic forgetting \cite{liang2024aide,liang2022balancing}.
We observe that the rapid adaptation to new modalities relies on the weight initialization of new experts and gates, and the insertion position of these new experts.
Thus, we introduce a simple yet effective scheme for selecting positions and weights of new experts based on evaluating distribution shifts. 
Utilizing the data from the new modality, we fine-tune the existing gate layers of the MoE model. 
Then, we infer the new modality data to the models before and after fine-tuning
and get the average gate probability distribution for all samples.
By comparing the degree of gate probability distributions before and after fine-tuning, we identify the top-k layers for adding experts by examining the magnitude of these shifts.
Then, based on the probability distribution after fine-tuning,
we determine the expert with the highest probability and replicate the gate and expert weights onto the newly incorporated expert.

% \gsh{Don't forget to add numbers in this part!} 
Experimental results show that MoExtend achieves a training speed acceleration $\sim$6 times faster than full fine-tuning, while also delivering superior performance. The positions selection scheme in MoExtend allows for fewer newly added experts, specifically, half the number of new experts required for the Mixtral model, which reduces training time to $\sim$30 hours without compromising performance. 
In addition, MoExtend helps mitigate the risk of catastrophic forgetting when extending MoE LLMs to handle multimodal inputs.
Our contributions can be summarized as follows:
\begin{itemize}
\vspace{-0.2cm}
\item We introduce MoExtend, a strategy designed to augment Mixture-of-Experts LLMs with new modalities by addition of new experts.
\vspace{-0.2cm}
\item MoExtend offers significant advantages, including substantially reduced fine-tuning costs, no additional costs during inference, and a minimized impact from catastrophic forgetting issue.
\end{itemize}

% \subsection{}

\section{Methodology}
\label{sec:method}
\begin{figure*}[t]
  \centering
 \includegraphics[width=0.9\linewidth]{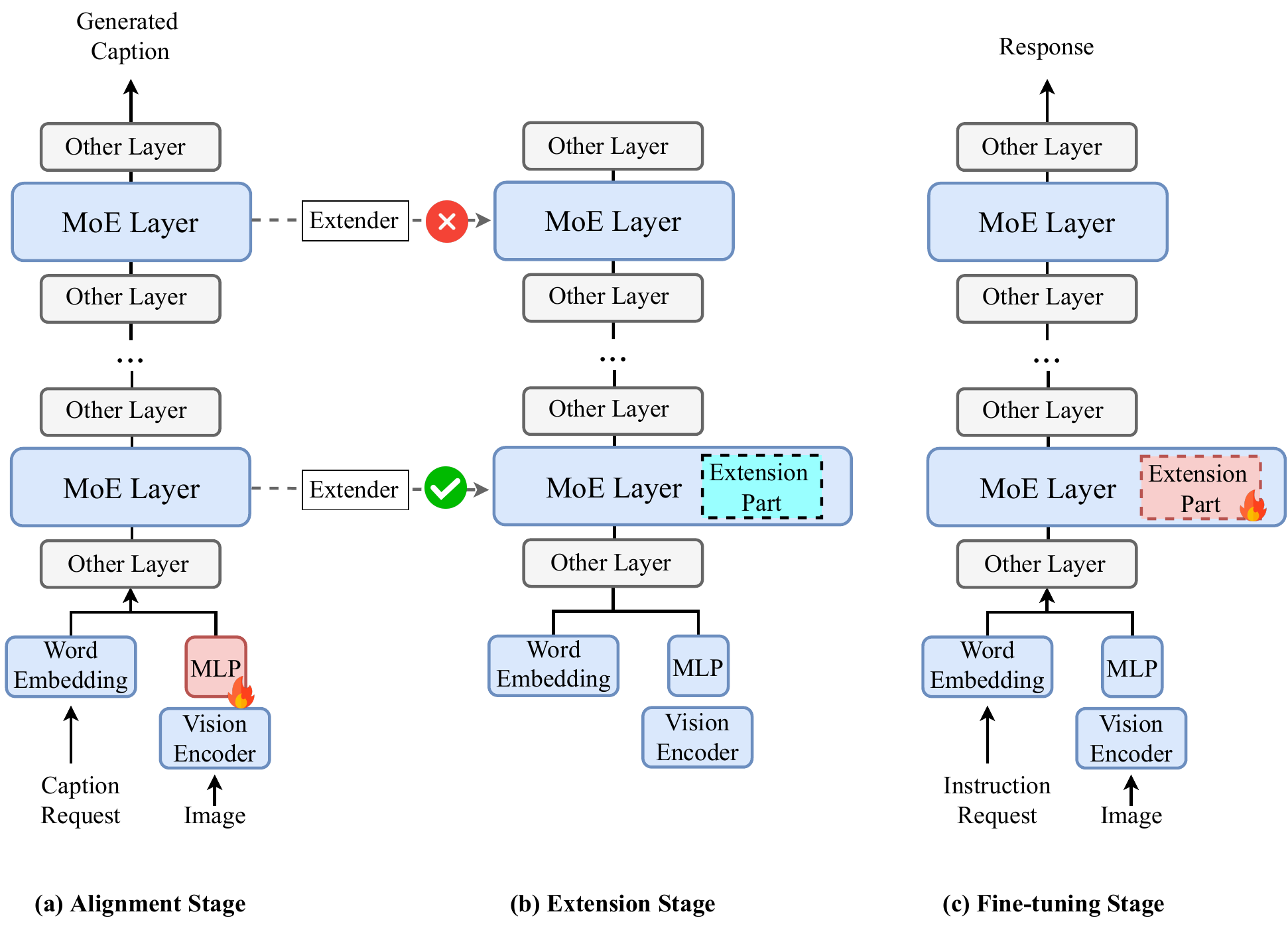}
  \caption{MoExtend consists of three stages: (a) Alignment Stage: we add a trainable MLP for pretrain vision encoder and tune the added MLP using image-caption data to achieve modal alignment; (b) Extension Stage: Determining which MoE layers need extension using an Extender; (c) Fine-tuning Stage: Fine-tuning the added extension part using a given Instruction dataset while keeping other parameters frozen. The "Other layer" represents other neural network components besides the MoE layer, including normalisation, self-attention layer, etc.}
\label{fig:zt}
\end{figure*}

In this section, we introduce MoExtend as an example of extending the visual modality for MoE models, which were originally designed for text modality only. As shown in Fig.~\ref{fig:zt}, MoExtend consists of three stages: alignment, extension with extender, and fine-tuning for the extension part. The purpose of the alignment stage is to initially align the MoE LLM with the newly added visual modality using a pre-trained vision encoder. The extension stage determines which MoE layers should be extended to accommodate the new modality information. The fine-tuning stage is then employed to tune the newly added parameters, achieving the final expansion of multimodal information.

% We provide a detailed explanation of the network architecture of MoExtend in Section \ref{sec:moextend}, and elaborate on the specifics of the extender and extension part in Sections \ref{sec:extender} and \ref{sec:part}, respectively.

\begin{figure*}[t]
  \centering
 \includegraphics[width=0.8\linewidth]{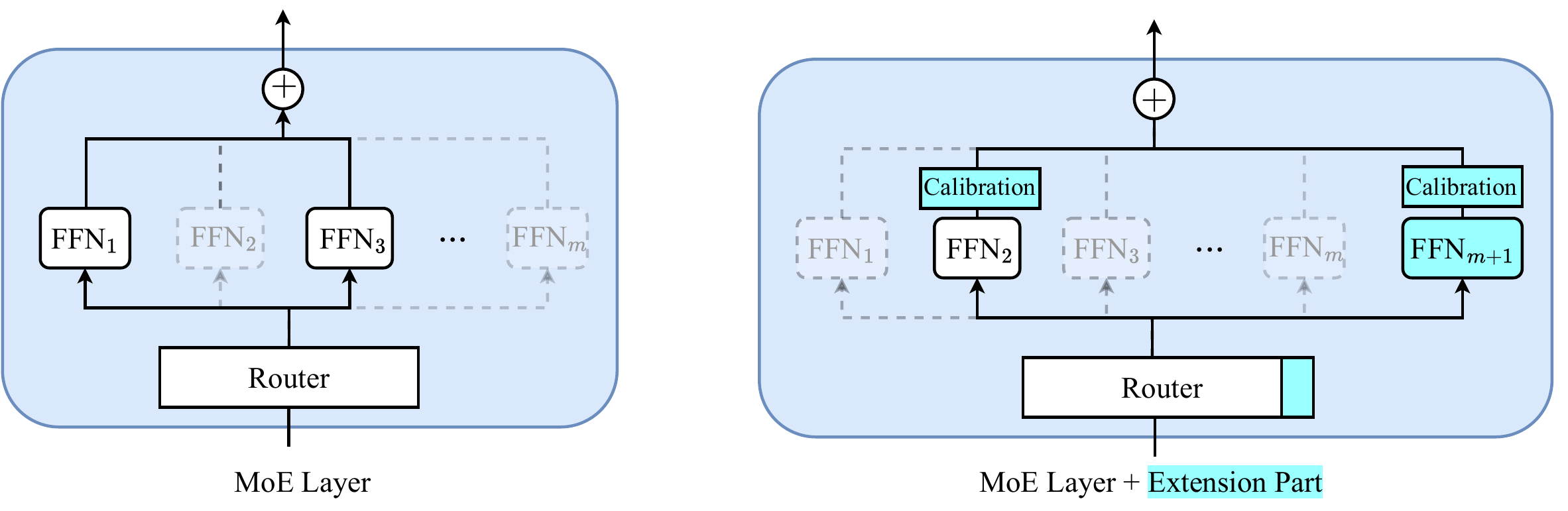}
  \caption{(Left) Original MoE layer; (Right) The extension part includes an additional expert FFN$_{m+1}$ and a corresponding column of trainable matrix parameters in the Router. Each expert is equipped with a learnable lightweight calibration module to correct gate weights altered due to the increased number of experts.}
\label{fig:extend}
\end{figure*}

\subsection{Alignment Stage}
\label{sec:stage1}

% \textbf{In the alignment stage}, 

As illustrated in Fig. \ref{fig:zt} (a), we train the newly added MLP using image-caption pairs from the LLaVA 1.5-558k dataset. This training aligns the modal information of images through the vision encoder (i.e., CLIP encoder) with textual modalities. Specifically, the caption $c$ from the textual modality is projected via word embedding to $T = [t_i]_{i=1}^N \in \mathbb{R}^{N\times D}$, where $D$ is the hidden size of LLM. Additionally, the image $I$ is mapped through the vision encoder to $V = [v_i]_{i=1}^P \in \mathbb{R}^{P\times D}$, where $P$ is the sequence length of visual tokens. Subsequently, the information from both modalities, $T$ and $V$, is concatenated into the vector $\mathbf{x}_0 \in \mathbb{R}^{(N+P)\times D}$.
For an $L$-layer MoE LLM, the forward process can be formulated as follows:
\begin{equation}
\begin{gathered}
\mathbf{x}_{\ell}^{\prime}=\operatorname{MSA}\left(\mathrm{LN}\left(\mathbf{x}_{\ell-1}\right)\right)+\mathbf{x}_{\ell-1}, \ell=1 \ldots L, \\
\mathbf{x}_{\ell}=\operatorname{MoE}\left(\operatorname{LN}\left(\mathbf{x}_{\ell}^{\prime}\right)\right)+\mathbf{x}_{\ell}^{\prime}, \ell=1 \ldots L,
\end{gathered}
\end{equation}
where MSA represents the multi-head self-attention module and LN represents layer normalization. The final input to the model is $\operatorname{LN}(\mathbf{x}_L)$.
During this stage, the structure of the MoE layer with $m$ experts remains unchanged, as depicted in Fig.~\ref{fig:extend} (Left). The router predicts the probability of each token being assigned to each expert, and each token is computed by the top-$k$ experts with the highest probabilities. The output of the MoE layer is a weighted sum as follows:
\begin{equation}
\operatorname{MoE}(\mathbf{x}) = \sum\nolimits_{j=1}^k s(\mathbf{x})_j\cdot \operatorname{FFN}(\mathbf{x})_j,
\label{dd}
\end{equation}
where $k \leq m$. Note that the weighted summation in Eq.~(\ref{dd}) is related to the outputs of experts with top-k probability. The parameter $k$ has a significant impact on MoE LLMs. However, to consider the trade-off between training efficiency and model performance, it's common to set $k=2$. In this paper, we also follow this setting. The $[\operatorname{FFN}_i]_{i=1}^m$ represents $m$ experts, and
\begin{equation}
s(\mathbf{x})_j = e^{f(\mathbf{x})_j} \large/ \sum\nolimits_{h = 1} ^m e^{f(\mathbf{x})_h},
\label{eq:softmax}
\end{equation}
where $f(\mathbf{x}) = \mathbf{W} \mathbf{x}$ and $\mathbf{W} \in \mathbb{R}^{D\times m}$ are the parameters of the router. 

% \vspace{0.4cm}
% \myPara{In the Extension stage}, as

\subsection{Extension Stage}
\label{sec:stage2}

To address the incorporation of additional modality information via extending the MoE layer, the most straightforward approach is to add a new expert to each MoE layer. However, this approach not only increases the parameter count significantly, leading to greater computational costs during training but also poses a potential risk of overfitting due to blindly adding a large number of parameters.

Therefore, in the extension stage, inspired by the concept of neural network pruning~\cite{li2016pruning,gao2020sod100k}, we construct an Extender to adaptively determine whether each MoE layer needs extension. Specifically, we randomly sample 10,000 instruction data related to the vision modality from the LLaVA 1.5-mix-665k dataset~\cite{liu2023llava} as the validation set $S_e$, with the remaining data forming the sub-training set $S_t$.

Next, for the model $\kappa$ obtained from the alignment stage training, we make all routers of the MoE layers trainable while freezing all other parameters. Utilizing $S_t$, we tune $\kappa$ for 1,000 steps to obtain $\kappa^\prime$. Furthermore, we input $S_e$ into both $\kappa$ and $\kappa^\prime$, and count the occurrences of each expert being selected in every MoE layer, resulting in 
\begin{equation}
\vspace{-0.2cm}
R_\kappa =     \{r^\kappa_{ij}\}_{m\times L}, \quad 
R_{\kappa^\prime} = \{r^{\kappa^\prime}_{ij}\}_{m\times L}.
\end{equation}
After normalization as follows, we can estimate the probability distributions of each expert being selected in every MoE layer:
% \begin{equation}
% \bar{R}_\kappa = R_\kappa \large/  \sum_{j=1}^L r^\kappa_{1j}, \quad \bar{R}_{\kappa^\prime} = R_{\kappa^\prime} \large/  \sum_{i=1}^L r^{\kappa^\prime}_{1j}.
% \end{equation}
\begin{equation}
\vspace{-0.2cm}
\begin{gathered}
\bar{R}_\kappa = R_\kappa \large/  (r^\kappa_{11} + r^\kappa_{21} + ...+ r^\kappa_{m1}), \\
\bar{R}_{\kappa^\prime} = R_{\kappa^\prime} \large/   (r^{\kappa^\prime}_{11} + r^{\kappa^\prime}_{21} + ...+ r^{\kappa^\prime}_{m1}).
\end{gathered}
\end{equation}
It is worth noting that for $1 \leq i \leq L$, $\sum_{i=1}^m r^\kappa_{i1} = \sum_{i=1}^m r^\kappa_{ij}$ and $\sum_{i=1}^m r^{\kappa^\prime}_{i1} = \sum_{i=1}^m r^{\kappa^\prime}_{ij}$.
Then, we can estimate the distribution differences of expert selections in each MoE layer between the two models $\kappa$ and $\kappa^\prime$ by calculating $d_j$ as follows:
% \begin{equation}
%     d_i = \operatorname{Std}(\bar{r}^\kappa_{i1}-\bar{r}^{\kappa^\prime}_{i1},\bar{r}^\kappa_{i2}-\bar{r}^{\kappa^\prime}_{i2},...,\bar{r}^\kappa_{iL}-\bar{r}^{\kappa^\prime}_{iL}  ), 1\leq i \leq m,
%     \label{eq:std}
% \end{equation}
\begin{equation}
\begin{gathered}
d_j = \operatorname{Std}_{i=1}^m(\bar{r}^\kappa_{ij}-\bar{r}^{\kappa^\prime}_{ij}), 1\leq j \leq L,
\label{eq:std}
\end{gathered}
\end{equation}

where $\operatorname{Std}$ denotes standard deviation. If $d_j$ is small, it implies that the MoE layer $j$ exhibits minimal response variation to the current data of the image-text modality, hence, there's no necessity to add new experts to this layer. Conversely, for MoE layers with larger $d_j$, adding new experts can effectively address the learning of new modality information.
We rank the MoE layers based on $d_j$ and introduce a new expert FFN$_{m+1}$ to the top $\lfloor pL \rfloor$ layers for original MoE LLM $\kappa$, with $p$ set to 0.5 in this paper. In fact, the adaptive extension stage proposed in this section not only reduces computational costs during training and mitigates the risk of overfitting but also accelerates the training of MoE LLM. For detailed analysis, please refer to Section \ref{sec:ana}.

\subsection{Fine-tuning Stage}
\label{sec:stage3}

In addition to introducing an additional expert in certain MoE layers for the original $\kappa$, as mentioned in Section \ref{sec:stage2}, and illustrated in Fig. \ref{fig:extend}, we also need to augment the parameters of the corresponding routers for these experts, i.e.,
\begin{equation}
\begin{gathered}
\mathbf{W}_{\text{new}} = [\mathbf{W};\mathbf{v}_{\text{new}}] \in \mathbb{R}^{D\times (m+1)}, 
\end{gathered}
\end{equation}
where $\mathbf{v}_{\text{new}} \in \mathbb{R}^{D\times 1}$, Furthermore, we add some Calibration modules to all experts in the MoE layers to mitigate changes in gate weights due to the addition of modalities. These newly introduced trainable parameters constitute the extension part. In this section, we fine-tune the extension part using the LLaVA 1.5-mix-665k dataset to enhance the final performance of LLM.

Specifically, we first consider the initialization of the newly added $m+1$-th expert and its corresponding router parameters $\mathbf{v}_{\text{new}}$. In this work, for the $j$-th MoE layer, we consider directly copying the expert and router parameters corresponding to
% \vspace{0.2cm}
\begin{equation}
\vspace{-0.2cm}
\max(r^\kappa_{1j} , r^\kappa_{2j} , \cdots , r^\kappa_{mj}),
\end{equation}
as initialization for the new parameters. This is because intuitively, the newly added expert is primarily intended to address the new modalities, and it is appropriate to initialize it with the existing expert that has the highest response to the new modalities. In Section \ref{sec:ana}, we will demonstrate that the initialization of the new parameters significantly affects the probability of an expert being selected by the MoE mechanism, thereby affecting the final performance of the MoE LLM.

Furthermore, since some MoE layers have added experts, $s(\mathbf{x})_j$ will change according to Eq.~(\ref{eq:softmax}). For example, for a fixed input $\mathbf{x}$, the new probability $s(\mathbf{x})_j^\prime$ satisfies
\begin{equation}
\vspace{-0.2cm}
\begin{gathered}
s(\mathbf{x})_j^\prime = e^{f(\mathbf{x})_j} \large/  (\sum\nolimits_{h = 1} ^m e^{f(\mathbf{x})_h} + e^{f(\mathbf{x})_{m+1}} )\\ \leq e^{f(\mathbf{x})_j} \large/ \sum\nolimits_{h = 1} ^m e^{f(\mathbf{x})_h} = s(\mathbf{x})_j,
\end{gathered}
\end{equation}
% \begin{equation}
%     s(\mathbf{x})_j^\prime = e^{f(\mathbf{x})_j} \large/  (\sum_{h = 1} ^m e^{f(\mathbf{x})_h} + e^{f(\mathbf{x})_{m+1}} ) \leq e^{f(\mathbf{x})_j} \large/ \sum_{h = 1} ^m e^{f(\mathbf{x})_h} = s(\mathbf{x})_j,
% \end{equation}
This causes the feature distribution of the original MoE $\kappa$ regarding previously learned knowledge to change during forward propagation, resulting in some degree of forgetting of existing knowledge by the model, thereby affecting performance. To address this issue, we add a Calibration module $s_c(\cdot)$ for each expert such that
\begin{equation}
\operatorname{MoE}(\mathbf{x}) = \sum\nolimits_{j=1}^k s(\mathbf{x})_j\cdot [1+s_c(\mathbf{x})] \cdot\operatorname{FFN}(\mathbf{x})_j,
\end{equation}
and $s_c(\cdot)$ is a two-layer GELU neural network $\mathbf{W}_1(\text{GELU}(\mathbf{W}_2(\cdot)))$. Here, the weights of $\mathbf{W}_1$ are initialized to 0, and $\mathbf{W}_2$ uses normal initialization. This initialization ensures that the calibration term $s_c(\mathbf{x}) = 0$, maintaining consistency with the model's output features when $s_c(\cdot)$ is not added, thus preventing significant interference with model output features due to the addition of $s_c(\cdot)$, which could lead to abnormal loss and affect model training.
For a fair comparison, all training hyperparameters, training methodologies, and loss functions with LLaVA 1.5-558k and LLaVA 1.5-mix-665k in all stages remain consistent with LLAVA.

% $\sum_{j=1}^L r^{\kappa^\prime}_{1j} = \sum_{j=1}^L r^{\kappa^\prime}_{ij}$

\section{Experiments}

\subsection{Experimental Setup}

% \vspace{-10pt}
\myPara{Model Settings.}
To ensure fairness in experimental comparisons, we follow the settings outlined in LLaVA 1.5. We utilize CLIP~\cite{radford2021learning} as the vision encoder, two linear layers with GELU~\cite{hendrycks2016gaussian} as the vision projection, and other training hyperparameters are shown in Appendix Table~\ref{tab:hyperparameters}.

\myPara{Dataset. }We utilize the same dataset as LLaVa 1.5 to train the model, consisting of  LLaVA 1.5-558k for pretraining stage and   LLaVA 1.5-mix-665k for instruction tuning stage~\cite{liu2023llava}. The computational cost of MoExtend is $\sim$15 hours of pretraining and $\sim$30 hours of visual instruction tuning, while MoExtend-Full, the model trained like LLaVA, need $\sim$200 hours of instruction tuning.

\begin{table*}[tbp]
  \centering
  \caption{Comparison with different LVLMs on 8 benchmarks. P, Res., PT, IT respectively represent parameters, the input image resolution, the number of samples in pretraining and instruction tuning stage. Evaluation benchmarks include two types: (1) image question answering: ScienceQA-IMG (SQA)~\cite{lu2022learn}, TextVQA (VQA$^{\text{T}}$)~\cite{singh2019towards}, VQA$^{\text{V2}}$~\cite{goyal2017making}; (2) benchmark toolkits: POPE~\cite{li2023evaluating}, MM-Vet~\cite{yu2023mm}, MMBench (MMB)~\cite{liu2023mmbench}, MMBench-Chinese (MMB$^{\text{CN}}$)~\cite{liu2023mmbench}, MME~\cite{fu2023mme}.  The best results and second best results are indicated by boldface and underline, respectively.}
  \resizebox*{\linewidth}{!}{
    \begin{tabular}{lcccc|ccc|ccccc}
    \toprule
    \multirow{2}[4]{*}{Model} & LLM   & \multirow{2}[4]{*}{Res.} & \multirow{2}[4]{*}{PT} & \multirow{2}[4]{*}{IT} & \multicolumn{3}{c|}{Image Question Answering} & \multicolumn{5}{c}{Benchmark Toolkit} \\
\cmidrule{6-13}          & Training \#P &       &       &       & SQA   & VQA$^\text{T}$ & VQA$^\text{V2}$ & POPE  & MM-Vet & MMB   & MMB$^\text{CN}$ & MME \\
    \midrule
    \rowcolor[rgb]{ .929,  .929,  .929} \textcolor[rgb]{ .502,  .502,  .502}{BLIP-2~\cite{li2023blip}} & \textcolor[rgb]{ .502,  .502,  .502}{13B} & \textcolor[rgb]{ .502,  .502,  .502}{224} & \textcolor[rgb]{ .502,  .502,  .502}{129M} & \textcolor[rgb]{ .502,  .502,  .502}{-} & \textcolor[rgb]{ .502,  .502,  .502}{61.0} & \textcolor[rgb]{ .502,  .502,  .502}{42.5} & \textcolor[rgb]{ .502,  .502,  .502}{41.0} & \textcolor[rgb]{ .502,  .502,  .502}{85.3} & \textcolor[rgb]{ .502,  .502,  .502}{22.4} & \textcolor[rgb]{ .502,  .502,  .502}{-} & \textcolor[rgb]{ .502,  .502,  .502}{-} & \textcolor[rgb]{ .502,  .502,  .502}{1293.8} \\
    \rowcolor[rgb]{ .929,  .929,  .929} \textcolor[rgb]{ .502,  .502,  .502}{InstructBLIP-7B~\cite{dai2023instructblip}} & \textcolor[rgb]{ .502,  .502,  .502}{7B} & \textcolor[rgb]{ .502,  .502,  .502}{224} & \textcolor[rgb]{ .502,  .502,  .502}{129M} & \textcolor[rgb]{ .502,  .502,  .502}{1.2M} & \textcolor[rgb]{ .502,  .502,  .502}{60.5} & \textcolor[rgb]{ .502,  .502,  .502}{50.1} & \textcolor[rgb]{ .502,  .502,  .502}{-} & \textcolor[rgb]{ .502,  .502,  .502}{-} & \textcolor[rgb]{ .502,  .502,  .502}{26.2} & \textcolor[rgb]{ .502,  .502,  .502}{36.0} & \textcolor[rgb]{ .502,  .502,  .502}{23.7} & \textcolor[rgb]{ .502,  .502,  .502}{-} \\
    \rowcolor[rgb]{ .929,  .929,  .929} \textcolor[rgb]{ .502,  .502,  .502}{InstructBLIP-13B~\cite{dai2023instructblip}} & \textcolor[rgb]{ .502,  .502,  .502}{13B} & \textcolor[rgb]{ .502,  .502,  .502}{224} & \textcolor[rgb]{ .502,  .502,  .502}{129M} & \textcolor[rgb]{ .502,  .502,  .502}{1.2M} & \textcolor[rgb]{ .502,  .502,  .502}{63.1} & \textcolor[rgb]{ .502,  .502,  .502}{50.7} & \textcolor[rgb]{ .502,  .502,  .502}{-} & \textcolor[rgb]{ .502,  .502,  .502}{78.9} & \textcolor[rgb]{ .502,  .502,  .502}{25.6} & \textcolor[rgb]{ .502,  .502,  .502}{-} & \textcolor[rgb]{ .502,  .502,  .502}{-} & \textcolor[rgb]{ .502,  .502,  .502}{1212.8} \\
    \rowcolor[rgb]{ .929,  .929,  .929} \textcolor[rgb]{ .502,  .502,  .502}{Shikra~\cite{chen2023shikra}} & \textcolor[rgb]{ .502,  .502,  .502}{13B} & \textcolor[rgb]{ .502,  .502,  .502}{224} & \textcolor[rgb]{ .502,  .502,  .502}{600K} & \textcolor[rgb]{ .502,  .502,  .502}{5.5M} & \textcolor[rgb]{ .502,  .502,  .502}{-} & \textcolor[rgb]{ .502,  .502,  .502}{-} & \textcolor[rgb]{ .502,  .502,  .502}{77.4} & \textcolor[rgb]{ .502,  .502,  .502}{-} & \textcolor[rgb]{ .502,  .502,  .502}{-} & \textcolor[rgb]{ .502,  .502,  .502}{58.8} & \textcolor[rgb]{ .502,  .502,  .502}{-} & \textcolor[rgb]{ .502,  .502,  .502}{-} \\
    \rowcolor[rgb]{ .929,  .929,  .929} \textcolor[rgb]{ .502,  .502,  .502}{IDEFICS-9B~\cite{laurenccon2024obelics}} & \textcolor[rgb]{ .502,  .502,  .502}{7B} & \textcolor[rgb]{ .502,  .502,  .502}{224} & \textcolor[rgb]{ .502,  .502,  .502}{353M} & \textcolor[rgb]{ .502,  .502,  .502}{1M} & \textcolor[rgb]{ .502,  .502,  .502}{-} & \textcolor[rgb]{ .502,  .502,  .502}{25.9} & \textcolor[rgb]{ .502,  .502,  .502}{50.9} & \textcolor[rgb]{ .502,  .502,  .502}{-} & \textcolor[rgb]{ .502,  .502,  .502}{-} & \textcolor[rgb]{ .502,  .502,  .502}{48.2} & \textcolor[rgb]{ .502,  .502,  .502}{25.2} & \textcolor[rgb]{ .502,  .502,  .502}{-} \\
    \rowcolor[rgb]{ .929,  .929,  .929} \textcolor[rgb]{ .502,  .502,  .502}{IDEFICS-80B~\cite{laurenccon2024obelics}} & \textcolor[rgb]{ .502,  .502,  .502}{65B} & \textcolor[rgb]{ .502,  .502,  .502}{224} & \textcolor[rgb]{ .502,  .502,  .502}{353M} & \textcolor[rgb]{ .502,  .502,  .502}{1M} & \textcolor[rgb]{ .502,  .502,  .502}{-} & \textcolor[rgb]{ .502,  .502,  .502}{30.9} & \textcolor[rgb]{ .502,  .502,  .502}{60.0} & \textcolor[rgb]{ .502,  .502,  .502}{-} & \textcolor[rgb]{ .502,  .502,  .502}{-} & \textcolor[rgb]{ .502,  .502,  .502}{54.5} & \textcolor[rgb]{ .502,  .502,  .502}{38.1} & \textcolor[rgb]{ .502,  .502,  .502}{-} \\
    \rowcolor[rgb]{ .929,  .929,  .929} \textcolor[rgb]{ .502,  .502,  .502}{Qwen-VL-7B~\cite{bai2023qwen}} & \textcolor[rgb]{ .502,  .502,  .502}{7B} & \textcolor[rgb]{ .502,  .502,  .502}{448} & \textcolor[rgb]{ .502,  .502,  .502}{1.4B} & \textcolor[rgb]{ .502,  .502,  .502}{50M} & \textcolor[rgb]{ .502,  .502,  .502}{67.1} & \textcolor[rgb]{ .502,  .502,  .502}{63.8} & \textcolor[rgb]{ .502,  .502,  .502}{78.8} & \textcolor[rgb]{ .502,  .502,  .502}{-} & \textcolor[rgb]{ .502,  .502,  .502}{-} & \textcolor[rgb]{ .502,  .502,  .502}{38.2} & \textcolor[rgb]{ .502,  .502,  .502}{7.4} & \textcolor[rgb]{ .502,  .502,  .502}{-} \\
    \rowcolor[rgb]{ .929,  .929,  .929} \textcolor[rgb]{ .502,  .502,  .502}{Qwen-VL-7B-Chat~\cite{bai2023qwen}} & \textcolor[rgb]{ .502,  .502,  .502}{7B} & \textcolor[rgb]{ .502,  .502,  .502}{448} & \textcolor[rgb]{ .502,  .502,  .502}{1.4B} & \textcolor[rgb]{ .502,  .502,  .502}{50M} & \textcolor[rgb]{ .502,  .502,  .502}{68.2} & \textcolor[rgb]{ .502,  .502,  .502}{61.5} & \textcolor[rgb]{ .502,  .502,  .502}{78.2} & \textcolor[rgb]{ .502,  .502,  .502}{-} & \textcolor[rgb]{ .502,  .502,  .502}{-} & \textcolor[rgb]{ .502,  .502,  .502}{60.6} & \textcolor[rgb]{ .502,  .502,  .502}{56.7} & \textcolor[rgb]{ .502,  .502,  .502}{1487.5} \\
    \rowcolor[rgb]{ .929,  .929,  .929} \textcolor[rgb]{ .502,  .502,  .502}{MoE-LLaVA-2.7B×4~\cite{lin2024moe}} & \textcolor[rgb]{ .502,  .502,  .502}{5B} & \textcolor[rgb]{ .502,  .502,  .502}{336} & \textcolor[rgb]{ .502,  .502,  .502}{558K} & \textcolor[rgb]{ .502,  .502,  .502}{1.6M} & \textcolor[rgb]{ .502,  .502,  .502}{68.5} & \textcolor[rgb]{ .502,  .502,  .502}{51.4} & \textcolor[rgb]{ .502,  .502,  .502}{77.6} & \textcolor[rgb]{ .502,  .502,  .502}{85.0} & \textcolor[rgb]{ .502,  .502,  .502}{34.3} & \textcolor[rgb]{ .502,  .502,  .502}{65.2} & \textcolor[rgb]{ .502,  .502,  .502}{-} & \textcolor[rgb]{ .502,  .502,  .502}{1335.1} \\
    \rowcolor[rgb]{ .929,  .929,  .929} \textcolor[rgb]{ .502,  .502,  .502}{MoE-LLaVA-2.7B×4~\cite{lin2024moe}} & \textcolor[rgb]{ .502,  .502,  .502}{5B} & \textcolor[rgb]{ .502,  .502,  .502}{384} & \textcolor[rgb]{ .502,  .502,  .502}{558K} & \textcolor[rgb]{ .502,  .502,  .502}{1.6M} & \textcolor[rgb]{ .502,  .502,  .502}{70.3} & \textcolor[rgb]{ .502,  .502,  .502}{57.0} & \textcolor[rgb]{ .502,  .502,  .502}{79.9} & \textcolor[rgb]{ .502,  .502,  .502}{85.7} & \textcolor[rgb]{ .502,  .502,  .502}{35.9} & \textcolor[rgb]{ .502,  .502,  .502}{68.0} & \textcolor[rgb]{ .502,  .502,  .502}{-} & \textcolor[rgb]{ .502,  .502,  .502}{1431.3} \\
    \rowcolor[rgb]{ .929,  .929,  .929} \textcolor[rgb]{ .502,  .502,  .502}{SPHINX-MoE~\cite{gao2024sphinx}} & \textcolor[rgb]{ .502,  .502,  .502}{8×7B} & \textcolor[rgb]{ .502,  .502,  .502}{448} & \multicolumn{2}{c|}{\textcolor[rgb]{ .502,  .502,  .502}{15.3M}} & \textcolor[rgb]{ .502,  .502,  .502}{74.5} & \textcolor[rgb]{ .502,  .502,  .502}{68.0} & \textcolor[rgb]{ .502,  .502,  .502}{81.1} & \textcolor[rgb]{ .502,  .502,  .502}{89.6} & \textcolor[rgb]{ .502,  .502,  .502}{40.9} & \textcolor[rgb]{ .502,  .502,  .502}{71.3} & \textcolor[rgb]{ .502,  .502,  .502}{-} & \textcolor[rgb]{ .502,  .502,  .502}{1485.3} \\
    \midrule
    LLaVA-1.5~\cite{liu2023improved} & 7B    & 336   & 558K  & 665K  & 66.8  & 58.2  & 78.5  & \underline{85.9}  & 30.5  & 64.3  & 58.3  & 1510.7 \\
    HyperLLaVA~\cite{canony2024hyperllava} & 7B    & 336   & 558K  & 665K  & 70.4  & 58.5  & \underline{79.1}  & \textbf{86.3}  & 31.0  & 65.9  & 60.6  & 1481.2 \\
    LLaVA-1.5~\cite{liu2023improved} & 13B   & 336   & 558K  & 665K  & \underline{71.6}  & \textbf{61.3}  & \textbf{80.0}  & \underline{85.9}  & 35.4  & \underline{67.7}  & \textbf{63.6}  & \underline{1531.3} \\
    MoExtend & 3B    & 336   & 558K  & 665K  & \textbf{73.8}  & \underline{58.7}  & 76.6  & 85.5  & 37.1  & \textbf{67.8 } & \underline{61.5}  & \textbf{1710.1} \\
    \bottomrule
    \end{tabular}%
  }
  \label{tab:multimodal}%
\end{table*}%

\subsection{Image Understanding Evaluation}

\myPara{Image Question Answering. }As shown in Table~\ref{tab:multimodal}, we assess MoExtend performance across four widely-used image question answering benchmarks. Compared to the state-of-the-art method LLaVA-1.5~\cite{liu2023llava}, MoExtend exhibits robust image understanding capabilities and achieves performance very close to that of LLaVA-1.5. Specifically, MoExtend, which is trained with only 3B LLM parameters, surpasses LLaVA-1.5 13B, trained with 13B LLM parameters, by 3.1\%, and outperforms the recent vision-language model HyperLLaVA~\cite{canony2024hyperllava} by over 4.8\% on SQA. Remarkably, MoExtend achieves comprehensive superiority over IDEFICS-80B~\cite{laurenccon2024obelics} with only 13B activated parameters, underscoring the strong comprehension abilities of MoE-LLaVA in vision features.

\myPara{Performance on Multimodal Benchmarks. }To comprehensively evaluate multimodal comprehension capabilities of MoExtend, we evaluate its performance across five widely-used benchmark toolkits, as shown in Table~\ref{tab:multimodal}. Experimental results indicate that, under the same dataset and training settings, MoExtend, fine-tuned with only 3B LLM parameters, achieves performance on par with the state-of-the-art model on most benchmark toolkits. Particularly, MoExtend has significantly superior performance on MME, surpassing the existing leading model LLaVA 1.5-13B by 178.8 points, indicating that MoExtend facilitates a efficient expansion of modalities.

\begin{table}[t]
  \centering
  \caption{Comparison on text benchmarks. We measure textual performance on a popular variety of tasks categorized as follow: (1) Commonsense Reasoning: ARC-Easy (Arc-e)~\cite{clark2018think}, Hellaswag (HellaS)~\cite{zellers2019hellaswag}, PIQA~\cite{bisk2020piqa}, Winogrande (WinoG)~\cite{sakaguchi2021winogrande}; (2) Code: MBPP~\cite{austin2021program}; (3) Popular aggregated results: MMLU~\cite{hendrycks2020measuring}; (4) Math: GSM8K~\cite{cobbe2021training}. MoExtend-Full is the model trained like LLaVA, which trains vision projection and LLM on instruction tuning stage. Avg. drop $\downarrow$ refers to the mean difference in performance metrics between the current model and its corresponding LLM. A smaller Avg. drop $\downarrow$ indicates less forgetting by the model and thus better performance. All evaluations are based on the open source toolkit OpenCompass.}
  \setlength{\tabcolsep}{2.5mm} % column spacing
  \resizebox*{\linewidth}{!}{
    \begin{tabular}{l|cccccccc}
    \toprule
    Model & Arc-e & HellaS & PIQA  & WinoG & MBPP  & MMLU  & GSM8K & Avg. drop $\downarrow$ \\
    \midrule
    Vicuna-7B~\cite{vicuna2023}  & 77.60  & 72.32  & 76.77  & 62.04  & 12.20  & 50.99  & 19.48 & - \\
    LLaVA-1.5-7B~\cite{liu2023llava}  & 80.07  & 72.02  & 76.22  & 62.51  & 15.00  & 51.61  & 19.64 & -0.81 \\
    \midrule
    Vicuna-13B~\cite{vicuna2023} & 85.36  & 75.67  & 78.45  & 65.75  & 25.20  & 56.67  & 29.66 & - \\
    LLaVA-1.5-13B~\cite{liu2023llava}  & 87.65  & 75.63  & 78.67  & 64.09  & 26.60  & 56.85  & 29.19 & -0.27 \\
    \midrule
    Phi2-2.7B~\cite{javaheripi2023phi} & 85.89 & 72.36 & 78.84 & 71.51 & 46.00 & 58.49 & 60.20 & - \\
    MoE-LLaVA-2.7B×4~\cite{lin2024moe}  & 87.30 & 70.83 & 79.38 & 69.61 & 10.00 & 47.92  & 53.22  & 7.86 \\
    \midrule
    Mixtral 8x7B~\cite{jiang2024mixtral}  & 92.24  & 81.84  & 81.61  & 70.48  & 36.40  & 71.17  & 71.95 & - \\
    MoExtend-Full & 88.36  & 77.40  & 80.63  & 64.56  & 34.80  & 69.02 & 67.83 & 3.30 \\
    MoExtend   & 93.12  & 80.75  & 81.50  & 69.69  & 34.60  & 71.12   & 72.03 & 0.41 \\
    \bottomrule
    \end{tabular}%
  }
  \vspace{-0.3cm}
  \label{tab:forgetting}%
\end{table}%

\myPara{Comparison with Forgetting. }To mitigate catastrophic forgetting in LVLMs, MoExtend fine-tunes LLM through calibration and the addition of new experts, thereby preserving the performance of LLM's original modalities. To evaluate the superiority of our fine-tuning strategy in preserving the understanding capabilities of LLM's original modalities, we evaluate the performance of LVLMs using different fine-tuning methods on pure text metrics as shown in Table~\ref{tab:forgetting}. Specifically, we compare the performance of LLaVA-1.5, MoExtend-Full, MoE-LLaVA, and MoExtend with original LLMs in Table A. Across all metrics, MoExtend exhibits performance similar to the original LLM. Additionally, we observe only slight decreases for LLaVA-1.5, while MoE-LLaVA and MoExtend-Full show significant declines relative to the original LLM model in pure text evaluation metrics, suggesting that full-parameter fine-tuning may lead to catastrophic forgetting for MoE-type LLMs, whereas non-MoE-type LLMs are less affected.

% \myPara{Computational Cost. }

\section{Ablation Study and Analysis}

\label{sec:ana}
\begin{table}[t]
  \centering
  \caption{Comparison of MoExtend with different architectures at 1k iterations. \#Layer represents the number of layers added expert. First-half indicates that new experts are only added to the first half layers of model, Second-half represents that only the second half layers of model have new experts, Interval means that we add new experts to every alternate layer of the model, First-quarter indicates only first quarter layers are
  added new expert, and First-interval means that we add new experts to first half layers alternately. }
  \resizebox*{1.\linewidth}{!}{
    \setlength{\tabcolsep}{5.5mm} % column spacing
    \begin{tabular}{lc|ccccc}
    \toprule
    Architecture & \#Layer & POPE  & MM-Vet & MMB   & \multicolumn{1}{l}{VQA$^{\text{T}}$} & Avg. \\
    \midrule
    All layer & 32    & 84.0  & 34.7  & \textbf{63.7} & \textbf{56.1} & 59.6\\
    \midrule
    First-half & 16    & 84.5  & 35.3  & 63.1  & 55.6 & 59.6 \\
    Second-half & 16    & 81.3  & 36.1  & 59.5  & 52.4 & 57.3\\
    Interval & 16    & 83.5  & 36.1  & \textbf{63.7} & 55.6 & 59.7 \\
    First-quarter & 8     & \textbf{85.4} & 35.4  & 61.3  & 54.6 & 59.2 \\
    First-interval & 8     & 83.6  & 34.8  & 62.7  & 54.3 & 58.9 \\
    \midrule
    Ours  & 16    & 84.3  & \textbf{36.4} & 63.1  & 55.7 & \textbf{59.9} \\
    \bottomrule
    \end{tabular}%
    }
  \label{tab:position}%
\end{table}%

\begin{figure*}[t]
  \centering
 \includegraphics[width=0.94\linewidth]{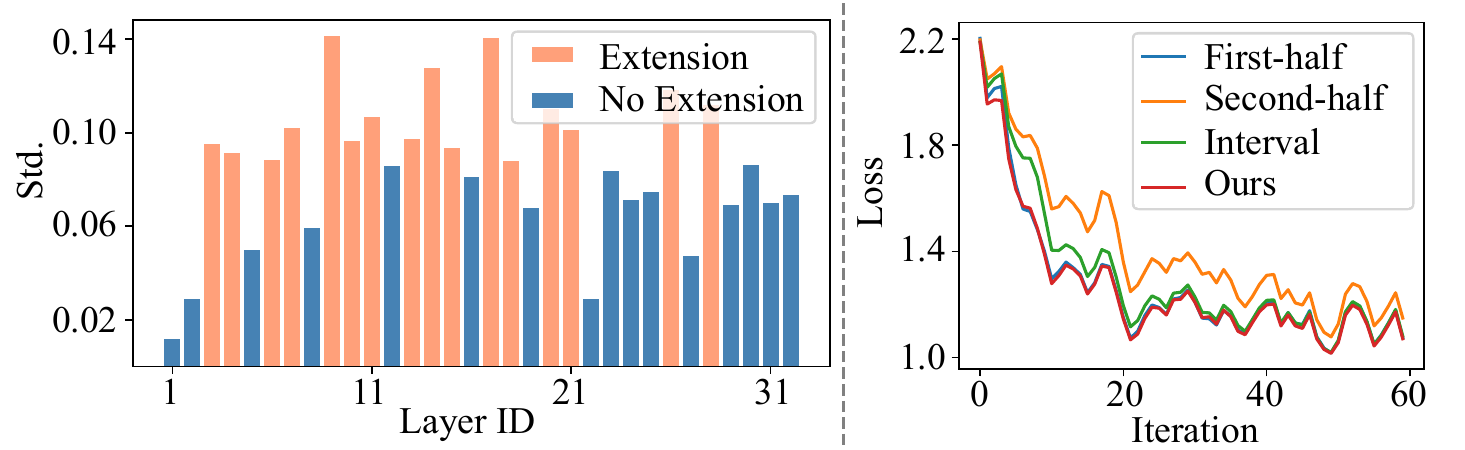}
  \caption{\textbf{Left}: std. $d_i$ of per layer caculated by Eq.~(\ref{eq:std}). Layers in orange color (layer id: 3, 4, 6, 7, 9, 10, 11, 13, 14, 15, 17, 18, 20, 21, 26, 28) are added new experts while layers in blue color are not with additional experts. \textbf{Right}: loss of MoExtend with by placing new expert layers in different positions. Employing our position selection scheme, we achieve faster convergence speeds compared to other manually designed schemes. }
\label{fig:loss}
\end{figure*}

\myPara{Effect of Model Architectures. }
We investigate the impact of different architectures on the performance of MoExtend. While the intuitive approach of adding new experts to all layers might seem optimal, our experiments, detailed in Table~\ref{tab:position}, reveal comparable performance between models with experts added to every layer (All layer), the first half (First-half), or every alternate layer (Interval). Additionally, results from models with experts added only to the first quarter (First-quarter) or every alternate layer starting from the first layer (First-interval) indicate performance degradation when too few layers receive additional experts. This finding informs our extension stage design, where experts are appropriately added to half of the layers.

As depicted in Fig.~\ref{fig:loss} (Left), our extension stage identifies layers requiring new experts. MoExtend based on our proposed strategy, as demonstrated in Table~\ref{tab:position}, performs on par with the current optimal insertion strategy (First-half, Interval). Furthermore, Fig.~\ref{fig:loss} (Right) shows that our extension strategy converges at a rate comparable to the optimal insertion strategy during training, validating its effectiveness on accurately determining the appropriate layers for adding new experts without extensive experimentation.

\begin{table}[t]
  \centering
  \caption{Comparison of MoExtend with different initial methods at 1k iterations. Copy($i$) means initializing new experts by copying the weight of original $i$-th expert.}
  \resizebox*{1.0\linewidth}{!}{
      \setlength{\tabcolsep}{7.5mm} % column spacing
    \begin{tabular}{ll|cccc}
    \toprule
    \multicolumn{2}{l|}{Method} & POPE  & MM-Vet & SQA   & VQA$^{\text{T}}$ \\
    \midrule
    \multirow{4}[2]{*}{Expert} & Copy(2) & 83.6  & 34.5  & 73.3  & 51.3 \\
          & Copy(4) & 83.7  & 35.1  & 71.7  & 54.6 \\
          & Copy(6) & 83.5  & 34.7  & 73.2  & 54.4 \\
          & Copy(8) & 83.7  & 34.7  & 74.1  & 54.8 \\
    \midrule
    \multirow{2}[2]{*}{Router} & Zero  & 83.6  & 34.8  & \textbf{74.4} & 54.8 \\
          & Mean  & 83.2  & 34.4  & 73.1  & 54.3 \\
    \midrule
    \multicolumn{2}{l|}{Ours} & \textbf{84.3} & \textbf{36.4} & 73.4  & \textbf{55.7} \\
    \bottomrule
    \end{tabular}%
    }
  \label{tab:init}%
  % \vspace{10pt}
\end{table}%

\myPara{Effect of Initialization. }
As depicted in Table~\ref{tab:init}, we analyze the impact of expert and router initialization on the performance of MoExtend. If the parameters of the new experts and router dimensions are directly copied from fixed positions $i$ of experts and corresponding dimensions of routers at each layer (Copy($i$)), the performance of copying experts from different positions is relatively close and lower than that of MoExtend. 

\begin{figure*}[t]
  \centering
 \includegraphics[width=0.94\linewidth]{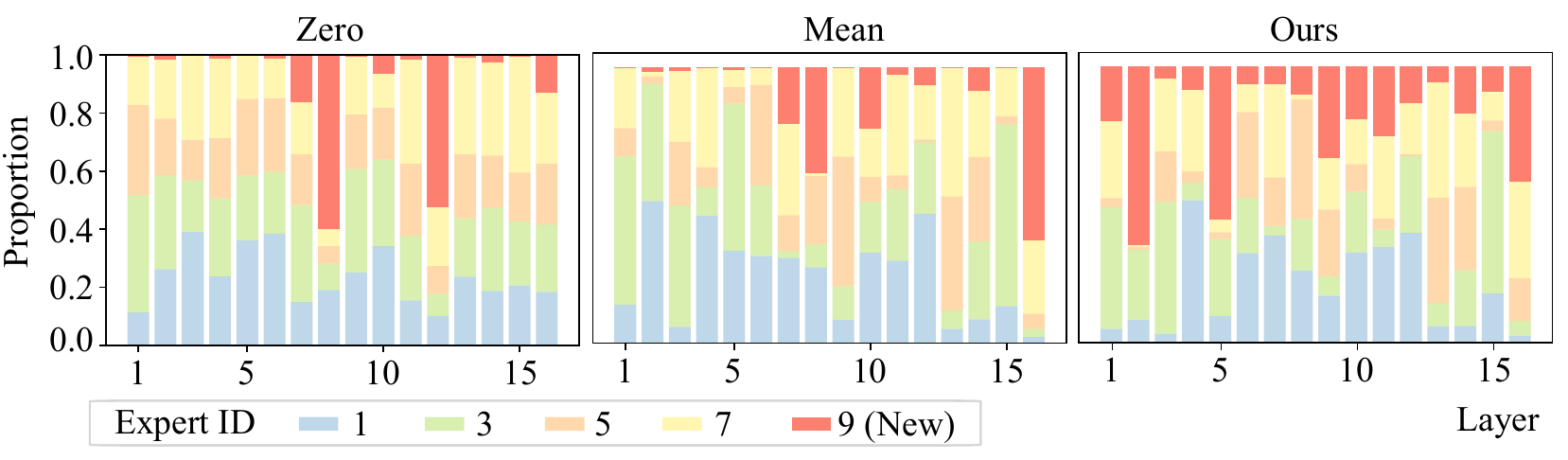}
  \caption{Distribution of expert selection per layer with different router initial methods. We randomly select 10,000 multimodal samples from  LLaVA 1.5-mix-665k as inputs and count the number of times each expert at each layer is selected. To streamline the visualization of results, we calculate and visualize the proportion of five experts. }
\label{fig:expert}
\end{figure*}
Additionally, we explore the performance when the router parameters are not directly copied from the corresponding router parameters of the $i$-th expert, but initialize directly with zeros or with the mean of the initial parameters of the eight experts (Mean). Experimental results indicate that initializing the router with zeros generally results in poorer performance compared to direct copying (Ours). Mean initialization implies that the new experts are a few selected in the initial state, and later in the instruction tuning stage the new experts are selected through gradient updates. In fact, this performance difference is mainly due to the fact that such an initialisation will lead to the newly added experts not being easily selected during the training process, so that the newly added experts are not fully trained or not used for new modality. Specifically, take the "Mean" initialisation as an example. Since the MoE layer generally selects the top-2 probability of experts for feature integration, the initialisation of "Mean" makes it difficult for the new experts to be selected with a large probability. Since the new router parameters and experts are rarely updated, it is difficult to improve this situation during the training process.

However, experimental results show that this initialization method leads to inferior performance. Furthermore, to investigate the impact of initialization methods on performance, we calculate the ratio of expert selection for different initializations as shown in Fig.~\ref{fig:expert}, and find that models initialized with Zero and Mean are both unbalanced in expert selection, while MoExtend is more balanced. This finding indicates that the balance of expert selection is closely related to model performance.

\begin{figure*}[t]
  \centering
 \includegraphics[width=0.94\linewidth]{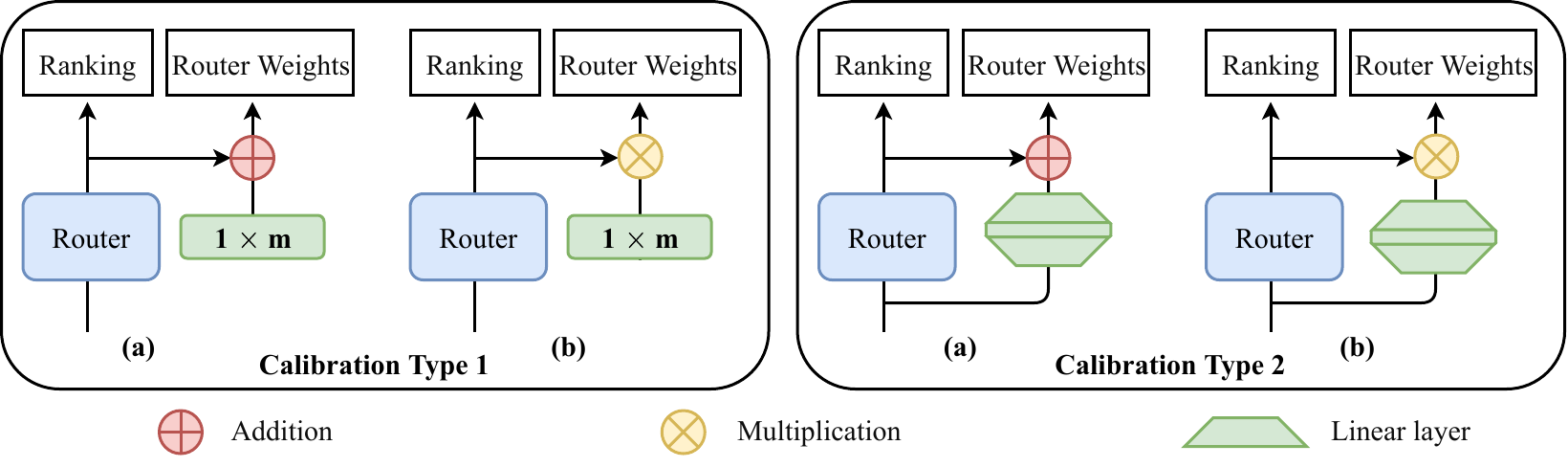}
  \caption{Structure of different types of calibration modules. The green modules represent calibration modules, and $m$ is the number of experts. The output of the calibration module acts on the softmax output of the router to correct the probability distribution effect caused by changes in the number of experts, ensuring proper gate weight adjustments for each expert. }
\label{fig:calib}
\vspace{-0.5cm}
\end{figure*}

\begin{table}[t]
  \centering
  \caption{Comparison of MoExtend with different calibration modules at 1000 iterations. The type of modules corresponds to Fig.~\ref{fig:calib}. The reason why Type2 (b) has no evaluation result is gradient explosion. "Zero" and "One" respectively denote filling all learnable parameters of the Calibration module with 0 or 1. "Zero+Normal" refers to initializing the two linear layers of the Calibration module in Type2 with 0 and standard normal values, respectively.}
   \resizebox*{1.0\linewidth}{!}{
    \setlength{\tabcolsep}{3.5mm} % column spacing
    \begin{tabular}{llccccc}
    \toprule
    Modules & Initialization & POPE  & MME   & SQA   & VQA$^{\text{T}}$ & Avg. \\
    \midrule
    Type1 (a) & Zero  & \textbf{84.8}  & 1495.2 & 72.4  & 53.2  & 426.4 \\
    Type1 (b) & One   & 83.5  & 1567.1 & 72.5  & \textbf{56.2}  & 444.8 \\
    Type2 (a) & Zero + Normal & 84.3  & \textbf{1571.0} & \textbf{73.4}  & 55.7  & \textbf{446.1} \\
    Type2 (b) & Normal + Normal & N/A     & N/A     & N/A    & N/A     & N/A\\
    \bottomrule
    \end{tabular}%
    }
  \label{tab:calibration}%
\vspace{-0.6cm}
\end{table}%

\myPara{The Design of Calibration Modules. }As shown in Fig.~\ref{fig:calib}, we design two concise calibration modules (Type1, Type2) to investigate the impact of these modules on MoExtend performance under two integration modes \cite{liang2020instance,huang2020dianet,zhong2023lsas,zhong2023esa}: addition (a) and multiplication (b). Type1 consists of a simple learnable parameter 1$\times$m, while Type2 consists of two simple linear layers connected by the GELU activation function. To minimize the disruption of router performance by calibration modules in the initial state, we mitigate the initial impact of calibration modules on routers through special initialization as shown in Table~\ref{tab:calibration}. In the additive mode of Type1, we use Zero initialization for calibration modules, while in the multiplicative mode, we use One initialization. 

In the additive mode of Type2, we initialize the first linear layer normally and zero-initialize the second linear layer. In the multiplicative mode, it is hard to reduce the impact of calibration modules through appropriate initialization, so we opt for simple normal initialization for both linear layers. Type2 (b) does not exhibit any evaluation result in Table~\ref{tab:calibration} because of gradient explosion, and the experimental results indicate that Type2 (a) calibration module structure performs better than others. 

%\hzz{limitation}
 
\section{Conclusion}
% \noindent{\textbf{Conclusion.}} 
In this work,  we introduce MoExtend, an effective framework tailored to streamline the modality adaptation and extension of Mixture-of-Experts (MoE) models. MoExtend introduces new experts into MoE models  by putting them at the parallel positions of  the experts in MoE. Then MoExtend designs a method to select previous experts in MoE for initilizing the new experts. Finally, it only tunes the new experts on the corresponding modal data and tasks.  This  endows  MoE with novel knowledge without necessitating the tuning of pretrained models such as MoE and vision encoders, thus avoiding the catastrophic forgetting issue. Furthermore,  MoExtend   facilitates rapid adaptation and extension to new modal data or tasks, thereby effectively addressing the challenge of accommodating new modalities within LLMs.  Empirical results show the efficacy and efficiency of MoExtend in augmenting the multimodal capabilities of LLMs. 

% \vspace{0.2cm}

\section{Limitation}
\label{sec:appendix}
% \noindent{\textbf{Limitation.}}
In this work, due to limited GPU resource, we take the visual task as one example to validate the effectiveness our proposed  MoExtend. So one limitation of
MoExtend is that its performance is not investigated on the other modal data, such as speech, and other tasks, e.g., continue learning and streaming  tasks.   
However, as aforementioned,  MoExtend is a general approach to extend the MoE model to other modal data or tasks, because our design principle is to  endows  MoE with novel knowledge via tuning the new integrated experts, and does not involve any specific tasks or modality. Accordingly, we believe that by replacing the vision encoder in MoExtend with other modal encoder and inserting new experts like MoExtend, one can easily extend MoExtend to other modal data and tasks, which is also left as our future work to thoroughly test.

\section{Related Work}

\subsection{Mixture of Experts}
Mixture of Experts (MoE)~\cite{masoudnia2014mixture,riquelme2021scaling,zhou2022mixture,lin2024moe,jiang2024mixtral} is a technique that leverages multiple sub-networks, also referred to as experts, to integrate features generated by different experts through adaptive strategies, thereby enhancing the overall performance of neural networks. The MoE layer, when processing each token, employs a router module to assign tokens to different experts, thereby reducing interference between different types of samples and keep low inference cost. 
In specific computational frameworks, MoE can achieve performance comparable to LLMs with a large amount of computational cost~\cite{masoudnia2014mixture}. Consequently, with the rapid advancement and application of LLMs, MoE is emerging as a promising and noteworthy paradigm for further enhancing LLM performance~\cite{masoudnia2014mixture,team2023gemini}. %\hzz{cite}
%moe 

\subsection{Multimodal Model}
Multimodal Learning involves leveraging various types of data, such as text, images, speech, and video, to train machine learning models for a more comprehensive understanding and inference capability~\cite{bayoudh2022survey,xu2023multimodal,zhong2023adapter,zhong2023let}. By integrating and jointly modeling different modalities of data, multimodal learning enhances machines' ability to comprehend and express rich real-world information, thereby improving performance in tasks like image description, sentiment analysis, speech recognition, and video understanding.
%多模态 moe 做扩展  我们的区别是什么

Recently, with the advancement of LLM technologies, multimodal learning methods have been rapidly integrated into LLM to expand its understanding and analysis of different modalities, especially visual modality~\cite{liu2023llava,bai2023qwen}. Recent efforts have focused on enhancing performance through methods such as adjusting datasets~\cite{liu2023llava}, optimizing training strategies~\cite{zhang2023llama,zhong2022cem}, improving image resolution~\cite{bai2023qwen}, enhancing image encoders~\cite{fan2024mousi,gao2024sphinx}, aligning inputs~\cite{radford2021learning}, and projecting layers~\cite{wu2023next,liu2023llava}. These approaches, by fine-tuning datasets and model scales through expanded visual instructions, have endowed LLM with robust visual comprehension capabilities. However, most current methods for expanding modalities generally involve fine-tuning a significant portion of or all parameters on multimodal data, leading to substantial computational costs and risking performance degradation due to forgetting. Facing this dilemma, in this paper, we consider leveraging the strong base performance of MoE  LLM to explore cost-effective methods for expanding LLM modalities by introducing new experts.

\section{Hyperparameters}
\begin{table}[h]
  \centering
  \vspace{-0.5cm}
  \caption{Training hyperparameters of MoExtend.}
      \setlength{\tabcolsep}{6.5mm} % column spacing
  \resizebox{0.99\hsize}{!}{
    \begin{tabular}{lcc}
    \toprule
    Hyperparameter & Pretrain & Fine-tune \\
    \midrule
    batch size & 256   & 128 \\
    learning rate & 1E-03 & 2E-05 \\
    schedule & cosine decay & cosine decay \\
    warmup ratio & 0.03  & 0.03 \\
    weight decay & 0     & 0 \\
    optimizer & AdamW & AdamW \\
    epoch & 1     & 1 \\
    aux loss coefficient & 0.001 & 0.001 \\
    precision & BF16  & BF16 \\
    GPU   & 8 × A800-80G & 8 × A800-80G \\
    text max length & 1024  & 2048 \\
    deepspeed stage & 2     & 3 \\
    \bottomrule
    \end{tabular}%
    }
  \label{tab:hyperparameters}%
\end{table}%

\section{Acknowledgments}

This work was supported by National Natural Science Foundation of China (No.61876045, 623B2099, U1711264). Pan Zhou acknowledges support from the Singapore Ministry of Education (MOE) Academic Research Fund (AcRF) Tier 1 grant.

\bibliography{custom}

\end{document}